\documentclass{article}

\usepackage{microtype}
\usepackage{graphicx, caption} 
\usepackage{subfigure}
\usepackage{booktabs} %
\usepackage{placeins}
\usepackage[utf8]{inputenc} %
\usepackage[T1]{fontenc}    %
\usepackage{url}            %
\usepackage{booktabs}       %
\usepackage{amsfonts}       %
\usepackage{nicefrac}       %
\usepackage{xcolor}         %
\usepackage{longtable}
\usepackage{amsmath}
\usepackage{wrapfig}
\usepackage{multirow}
\usepackage{fancyvrb,fvextra}
\usepackage{xspace}

\usepackage{hyperref}

\usepackage[arxiv]{icml2025} %

\usepackage{amsmath}
\usepackage{amssymb}
\usepackage{mathtools}
\usepackage{amsthm}

\usepackage[capitalize,noabbrev]{cleveref}

\theoremstyle{plain}

\theoremstyle{definition}

\theoremstyle{remark}

\usepackage[textsize=tiny]{todonotes}

\icmltitlerunning{SmolLM2}

\begin{document}

\twocolumn[
\icmltitle{SmolLM2: When Smol Goes Big — \\ Data-Centric Training of a Small Language Model}

\icmlsetsymbol{equal}{*}

\icmlsetsymbol{newline}{\\}
\begin{icmlauthorlist}
\icmlauthor{Loubna Ben Allal}{equal}
\icmlauthor{Anton Lozhkov}{equal}
\icmlauthor{Elie Bakouch}{equal}
\icmlauthor{Gabriel Martín Blázquez}{equal}
\icmlauthor{Guilherme Penedo}{}
\icmlauthor{Lewis Tunstall}{}
\icmlauthor{Andrés Marafioti}{}
\icmlauthor{Hynek Kydlíček}{}
\icmlauthor{Agustín Piqueres Lajarín}{}
\icmlauthor{Vaibhav Srivastav}{}
\icmlauthor{Joshua Lochner}{}
\icmlauthor{Caleb Fahlgren}{}
\icmlauthor{Xuan-Son Nguyen}{}
\icmlauthor{Clémentine Fourrier}{}
\icmlauthor{Ben Burtenshaw}{}
\newline
\icmlauthor{\hspace*{2cm}Hugo Larcher}{}
\icmlauthor{Haojun Zhao}{} 
\icmlauthor{Cyril Zakka}{}  
\icmlauthor{Mathieu Morlon}{}
\newline 
\icmlauthor{\hspace*{0.4cm}Colin Raffel}{}  
\icmlauthor{Leandro von Werra}{}
\icmlauthor{Thomas Wolf}{}
\end{icmlauthorlist}

\icmlkeywords{Machine Learning, ICML}

\icmlcorrespondingauthor{Loubna Ben Allal}{loubna@hf.co}
\icmlcorrespondingauthor{Leandro von Werra}{leandro@hf.co}
\icmlcorrespondingauthor{Thomas Wolf}{thomas@hf.co}

\begin{center}
\raisebox{-2pt}{\includegraphics[height=1.05em]{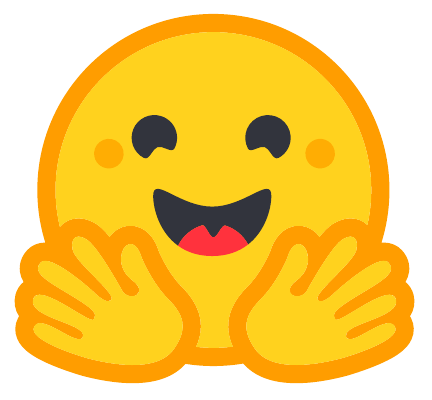}}\xspace Hugging Face
\vspace{1em}
\url{https://hf.co/collections/HuggingFaceTB/smollm2-6723884218bcda64b34d7db9}
\end{center}

]  %

\printAffiliationsAndNotice{\icmlEqualContribution} %

\begin{abstract}
While large language models have facilitated breakthroughs in many applications of artificial intelligence, their inherent largeness makes them computationally expensive and challenging to deploy in resource-constrained settings.
In this paper, we document the development of SmolLM2, a state-of-the-art ``small'' (1.7 billion parameter) language model (LM).
To attain strong performance, we overtrain SmolLM2 on \char`\~11 trillion tokens of data using a multi-stage training process that mixes web text with specialized math, code, and instruction-following data.
We additionally introduce new specialized datasets (FineMath, Stack-Edu, and SmolTalk) at stages where we found existing datasets to be problematically small or low-quality.
To inform our design decisions, we perform both small-scale ablations as well as a manual refinement process that updates the dataset mixing rates at each stage based on the performance at the previous stage.
Ultimately, we demonstrate that SmolLM2 outperforms other recent small LMs including Qwen2.5-1.5B and Llama3.2-1B.
To facilitate future research on LM development as well as applications of small LMs, we release both SmolLM2 as well as all of the datasets we prepared in the course of this project.
\end{abstract}

\vspace{-0.5cm} %
\section{Introduction}
\vspace{-0.1cm} %

Large language models (LMs) have become a cornerstone of modern AI systems due to their ability to follow natural language instructions and flexibly perform a huge range of tasks \cite{touvron2023llama,bai2023qwen,brown2020language,dubey2024llama,groeneveld2024olmo,chowdhery2023palm,young2024yi,taylor2022galactica}.
LLMs are, by their nature, \textit{large}, in the sense that they are models with many parameters (more than \char`\~10 billion, by current conventions).
This enormity results in enormous computational costs, both during training and for inference, which can prevent LLMs from being used in resource-constrained settings.
To address this issue, a flurry of recent work has aimed to produce performant \textit{small} (\char`\~3 billion parameters or less) LMs \cite{gunter2024apple,Yang2024Qwen2TR,llama3_2modelcard,team2024gemma,li2023textbooks}.
These small LMs are computationally inexpensive and can be run on a wider range of devices (e.g.\ mobile phones) while providing satisfactory performance on many important tasks.

A key factor in the performance and behavior of LMs is the data used to train them.
While important for an LM of any size, data curation has an especially outsized influence for smaller models, as their limited capacity must be carefully optimized for learning core knowledge and fundamental capabilities rather than memorizing incidental facts~\cite{abdin2024phi3,rolnick1705deep}. 
Most LMs are primarily trained on text crawled from the web \cite{radford2019language,raffel2020exploring} and state-of-the-art pipelines include sophisticated filtering and processing stages that aim to improve data quality \cite{li2024datacomp,penedo2024refinedweb,Penedo2024TheFD,soldaini2024dolma}.
Recently, it has become common to include ``specialized'' data from certain domains such as software code \cite{kocetkov2022stack,lozhkov2024starcoder} and mathematics \cite{paster2023openwebmath,han2024infimm}, which can improve performance not only on those domains but also more generally on challenging tasks that require reasoning \cite{muennighoff2023scaling,aryabumi2024code}.

Motivated by the above considerations, our contributions in this paper are as follows:
First, we perform a careful evaluation of existing web, code, math, and instruction-following datasets (\cref{sec:datasets}) to help guide training data design choices, ultimately training SmolLM2 via a multi-stage manual rebalancing of different sources to maximize performance (\cref{sec:pretraining-phases}).
Such on-the-fly rebalancing is a promising approach for large-scale training runs which can be sufficiently costly (around 1e23 FLOPs, or \$250,000 USD worth of GPU compute for SmolLM2) to preclude running multiple full-scale training runs.
Following standard practice, we also develop an instruction-tuned variant of SmolLM2 (\cref{sec:instruction-tuning}).
Additionally, after finding that existing datasets were too small and/or low-quality, we created the new datasets FineMath, Stack-Edu, and SmolTalk (for mathematics, code, and instruction-following respectively).
Ultimately, we show that both the base and instruction-tuned variants of SmolLM2 are state-of-the-art among similarly sized models (\cref{sec:evaluation-basemodels} and \cref{sec:evaluation_smollm2}).

\vspace{-0.25cm} %
\section{Background}
\vspace{-0.1cm} %
\label{sec:background}
Training a modern LM typically begins with ``pretraining'' on a large amount (e.g.\ trillions of tokens) of unstructured text.
Pretraining helps the model fit the structure of language \cite{clark2019does} and store factual knowledge \cite{petroni2019language,roberts2020much} and therefore has proven to be a vital part of LM training, which made the composition of the pretraining dataset a key consideration.
The data-hungry nature of pretraining has led to the use of large-scale web scrapes~\cite{commoncrawl,openai_gptbot,anthropic_claudebot} which in their raw form can lead to poorly performing LMs \cite{penedo2024refinedweb}.
Consequently, the primary means of curation for modern LM pretraining datasets involves designing sophisticated pipelines for automatically filtering and reformatting web texts~\cite{Penedo2024TheFD,penedo2024refinedweb,soldaini2024dolma,cerebras2023slimpajama,li2024datacomp} that aim to keep enough data to avoid detrimental repetition \cite{muennighoff2023scaling} while discarding any data that is not ``high-quality''.

Apart from web text, including ``specialized'' data from certain domains -- code \cite{kocetkov2022stack,li2023starcoder} and math \cite{paster2023openwebmath,han2024infimm,wangmathpile,azerbayev2023llemma} in particular -- can improve model performance on tasks that involve reasoning and world knowledge~\cite{muennighoff2023scaling,aryabumi2024code,lewkowycz2022solving,shao2024deepseekmath}.
The contribution of small specialized datasets can be dwarfed by much larger web-based pretraining data sources, which has led to the design of multi-stage pretraining where specialized or high-quality datasets are incorporated later in training \cite{abdin2024phi,olmo2_2024,blakeney2024does,singer2024h2o}.

After pretraining, language models typically undergo two additional training stages before deployment: instruction tuning and preference learning.
In instruction tuning, the model undergoes supervised training on instruction/response pairs that reflect the way that the language model should answer a query~\cite{wei2021finetuned,mishra2021cross,sanh2021multitask,wang2022self}.
This process provides a valuable way of tailoring LMs to provide helpful responses rather than simply attempting to continue the input (as taught during pretraining).
During preference learning, language models are further ``aligned'' towards their intended use by being trained to distinguish between helpful and unhelpful responses \cite{ouyang2022training,bai2022constitutional}.
This final stage typically involves a form of reinforcement learning \cite{bai2022constitutional,leerlaif,rafailov2024direct} on data labeled with human or synthetically generated preferences.

\section{Pretraining datasets}
\label{sec:datasets}
Pretraining data curation is especially important for small LMs due to their tendency to be more sensitive to noise in the training data \cite{rolnick1705deep,abdin2024phi3}.
In addition, designing a pretraining strategy involves not only selecting and curating data, but also determining how much to ``mix'' (i.e.\ sample) from different sources, which can be particularly important when including e.g.\ specialized math and code datasets.
We therefore undertook a careful evaluation of existing datasets and, wherever we deemed necessary, created new, improved, and larger datasets.

\subsection{Ablation setup}
\label{sec:ablation_setup}
To compare English web datasets and find the best mixture for training our models, we followed an empirical approach similar to \citet{Penedo2024TheFD}. Specifically, we trained models on each dataset under identical conditions: model configuration, training hyperparameters, and token count. We trained 1.7B parameter Transformers \citep{NIPS2017_3f5ee243} based on the Llama architecture~\cite{touvron2023llama}, with a sequence length of 2048, a global batch size of approximately 2 million tokens, the GPT-2 tokenizer~\cite{radford2019language}, and a cosine learning rate schedule \citep{loshchilov2016sgdr} with a learning rate of $3.0 \times 10^{-4}$.
Each dataset ablation model is trained on 350B tokens randomly sampled from the full dataset.
For evaluation, we also followed \citet{Penedo2024TheFD}, and used \href{https://github.com/huggingface/lighteval/}{\texttt{lighteval}} to evaluate on a variety of knowledge, reasoning, and text understanding benchmarks: MMLU~\cite{hendrycks2021measuring}, HellaSwag~\cite{zellers-etal-2019-hellaswag}, OpenBook QA~\cite{OpenBookQA2018}, PIQA~\cite{bisk2019piqa}, WinoGrande~\cite{sakaguchi2019winogrande}, ARC~\cite{clark2018think}, and CommonSenseQA~\cite{talmor-etal-2019-commonsenseqa}.

Math and code capabilities typically emerge only after extensive training, so  
similarly to~\citet{blakeney2024does,dubey2024llama,olmo2_2024}, when evaluating math and code datasets we started from a mid-training checkpoint of SmolLM2 at 3T tokens (detailed in \cref{sec:pretraining-phases}), which was trained primarily on web data. We then used an \textit{annealing} approach: the learning rate linearly decays to 0 while training on a mixture that includes the dataset under evaluation.
For math, we annealed on a mixture of 60B tokens of the dataset under evaluation and 40B from the pre-checkpoint mixture.
For code ablations, we performed annealing on 200B tokens, uniformly distributed across 15 of the most commonly used programming languages (\char`\~14B tokens each). 
We evaluated the math ablation models on GSM8K~\cite{cobbe2021training}, MATH~\cite{hendrycks2021measuring} and \href{https://huggingface.co/datasets/TIGER-Lab/MMLU-STEM}{MMLU-STEM} to assess their math capabilities using \texttt{lighteval}, and we used HumanEval~\cite{chen2021evaluating} and MultiPL-E~\cite{cassano2022multipl} to evaluate the code ablation models using the \href{https://github.com/bigcode-project/bigcode-evaluation-harness}{\texttt{BigCode-Evaluation-Harness}}.

\subsection{English web data}
Web text from Common Crawl has remained a popular source of pretraining data, and recent classifier-based filtering techniques have significantly advanced pretraining data quality~\cite{dubey2024llama,abdin2024phi,abdin2024phi3,kong2024large}.
Two prominent examples of open datasets that use classifer-based filtering are FineWeb-Edu~\cite{Penedo2024TheFD} and DCLM~\cite{li2024datacomp}.
FineWeb-Edu consists of 1.3T tokens that were deemed ``educational'' by a classifier trained on annotations generated by Llama3-70B-Instruct~\cite{dubey2024llama}.
DCLM comprises 3.8T tokens filtered using a fastText classifier~\cite{joulin2016fasttext,joulin2016bag} trained on instruction-following data from OpenHermes 2.5~\cite{OpenHermes-2.5} and high-scoring posts from the r/ExplainLikeImFive (ELI5) subreddit.
Training ablation models on 350B tokens each from FineWeb-Edu and DCLM attained performance shown in~\cref{tab:web-data-ablations}. 
We find that FineWeb-Edu achieves higher scores on the educational benchmarks MMLU, ARC, and OpenBookQA, while DCLM performs better on HellaSwag and CommonsenseQA.
These results align with the datasets' content: FineWeb-Edu prioritizes educational material, while DCLM captures more diverse, conversational styles.

\begin{table}[t]
\caption{Evaluation of models trained on FineWeb-Edu and DCLM for 350B tokens. 40/60 and 60/40 denote the FW-Edu/DCLM ratio.}
\vspace{0.5em}
\label{tab:web-data-ablations}
\centering
\small
\begin{tabular}{@{}lcccc@{}}
\toprule
\textbf{Task} & \textbf{FW-Edu} & \textbf{DCLM} & \textbf{40/60} & \textbf{60/40}  \\
\midrule
MMLU            & \textbf{37.5} & 35.5 & 36.5 & 37.0 \\
ARC             & \textbf{57.5} & 53.5 & 53.2 & 56.0 \\
OpenBookQA      & \textbf{41.9} & 40.8 & 39.0 & \textbf{41.9} \\
HellaSwag       & 60.1 & \textbf{62.3} & 61.4 & 62.2 \\
CommonsenseQA   & 36.2 & \textbf{40.1} & 39.9 & 38.5 \\
PIQA            & 76.2 & \textbf{76.9} & 75.7 & 76.4 \\
\bottomrule
\end{tabular}
\vspace{-1em}
\end{table}

Given the complementary strengths of FineWeb-Edu and DCLM, we explored whether mixing them could further improve performance. After testing different ratios, we found that a 60\% FineWeb-Edu and 40\% DCLM mix works well, as shown in~\cref{tab:web-data-ablations}: It nearly matches FineWeb-Edu's performance on MMLU, ARC, and OpenBookQA while also aligning with DCLM's results on HellaSwag and approaching its performance on CommonSenseQA.
Combining these datasets yields 5.1T tokens of (English) text.

\subsection{Math data}
\label{sec:math-datasets}
Specialized math pretraining data is crucial for developing robust mathematical understanding.
Recent research has shown that carefully curated mathematical content from Common Crawl, combined with targeted filtering techniques, can significantly enhance language models' mathematical reasoning capabilities~\cite{dubey2024llama,yang2024qwen2,shao2024deepseekmath,han2024infimm}.

\subsubsection{Comparison of Existing Datasets}
We compare two leading publicly available math datasets: OpenWebMath (OWM)~\cite{paster2023openwebmath} and InfiMM-WebMath~\cite{han2024infimm}.
OWM consists of 12B tokens, built by filtering math-specific content from Common Crawl and using a specialized text extraction pipeline to preserve mathematical formatting and equations.
InfiMM-WebMath contains 40B text tokens, and its authors show that it matches the performance of the private dataset of DeepSeekMath~\citep{shao2024deepseekmath}.

We performed annealing ablations (following the setup described in \cref{sec:ablation_setup}) on OWM and InfiMM-WebMath, finding that InfiMM-WebMath achieves a peak accuracy of 14\% on GSM8K compared to OWM’s 10\%, while OWM slightly outperforms InfiMM-WebMath on MATH. The full evaluation curves are available in~\cref{app:owm-vs-inf}. Despite training on 60B math tokens (i.e., 5 epochs for OWM and 1.5 epochs for InfiMM-WebMath), performance still lagged behind proprietary state-of-the-art small models~\cite{Yang2024Qwen2TR}. Further analysis highlighted two key limitations: insufficient dataset sizes, and insufficient focus on step-by-step mathematical reasoning, along with an overrepresentation of academic papers that focus on advanced concepts.

\subsubsection{New dataset: FineMath}

The aforementioned issues with OWM and InfiMM-WebMath motivated us to develop FineMath\footnote{\href{https://huggingface.co/datasets/HuggingFaceTB/finemath}{https://huggingface.co/datasets/HuggingFaceTB/finemath}}, a collection of up to 54B tokens of math data focusing on mathematical deduction and reasoning through classifier-based filtering.

We began by extracting text from Common Crawl WARC files using \href{https://resiliparse.chatnoir.eu/}{Resiliparse}, focusing on all 5.8B unique URLs from the FineWeb dataset (a subset of Common Crawl's 75B unique URLs).
We then employed the FineWeb-Edu filtering approach, using Llama-3.1-70B-Instruct~\citep{dubey2024llama} with a prompt (\cref{app:finemath-silver-prompt}) that scores content on a 3-point scale, where 1 indicates some mathematical content and 3 indicates step-by-step problem solutions at an appropriate level.
After training a classifier on these silver labels, we identified domains containing at least 10 pages with a quality score of 2 or higher.
We expanded our domain coverage by including domains with at least 10 URLs from either OWM or InfiMM-WebMath. From the Common Crawl index, we retrieved a total of 7.7B URLs belonging to this list of domains: 5.7B identified by our classifier, 0.6B from OWM, and 1.3B from InfiWebMath. 
We then re-extracted all identified pages using the OWM pipeline, preserving LaTeX formatting and removing all-boilerplate pages, yielding 7.1B pages containing 6.5T tokens.

To retain only high-quality math content, we reapplied a classifier trained on Llama-3.1-70B-Instruct annotations using a 5-point scale prompt (\cref{app:finemath-gold-prompt}) specifically targeting pages with reasoning and middle- to high-school-level content.
We note that InfiMM-WebMath used a similar classifier filtering pipeline, but their prompt did not target the same type of content.
After classification, we performed deduplication using single-band MinHash LSH~\cite{Broder1997MinHash} with 10 hashes and applied fastText language classification~\cite{joulin2016fasttext,joulin2016bag} to retain only English content.

Ultimately, we developed multiple variants of FineMath, including FineMath4+ (10B tokens, 6.7M documents) which retains only samples with scores of 4-5 and FineMath3+ (34B tokens, 21.4M documents) which includes scores 3-5. We additionally applied the same classifier to InfiMM-WebMath, creating Infi-WebMath4+ (8.5B tokens, 6.3M documents) and Infi-WebMath3+ (20.5B tokens, 13.9M documents). Similarly to~\citet{yang2024qwen2}, we decontaminate each dataset against GSM8K, MATH and MMLU using 13-gram matching and a minimum overlap ratio with the longest common subsequence of 0.6.

\vspace{-0.25cm} %

\begin{figure}
    \centering
    \includegraphics[width=1\linewidth]{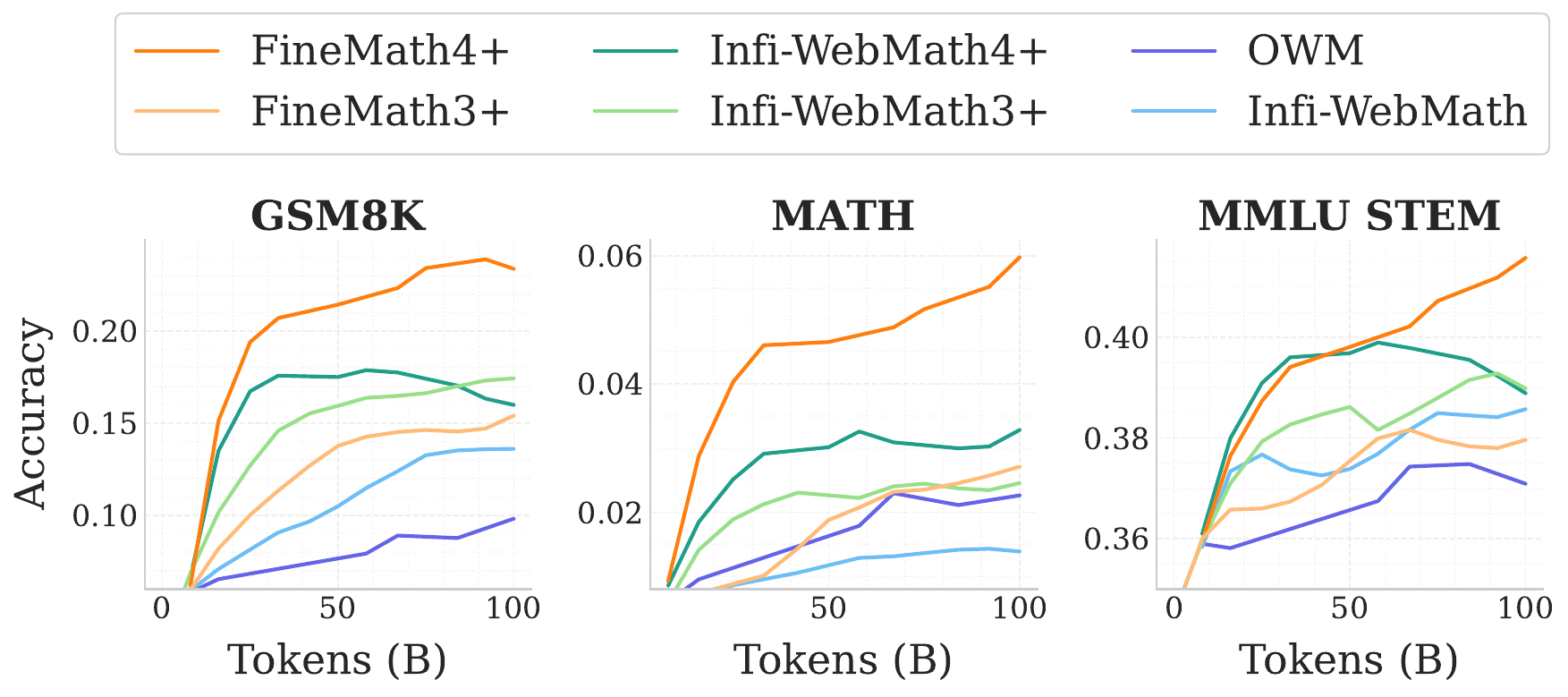}
    \caption{Performance of models trained on different subsets of FineMath and other math datasets.} %
    \label{fig:math_ablations}
    \vspace{-0.5cm} %
    
\end{figure}

\paragraph{Results} \cref{fig:math_ablations} presents our FineMath annealing ablations. All FineMath subsets consistently outperform OWM and InfiMM-WebMath on GSM8K, MATH, and MMLU-STEM. FineMath4+ achieves a 2x improvement on GSM8K and a 6x improvement on MATH compared to InfiMM-WebMath, demonstrating the importance of retaining high-quality mathematical content with reasoning. 
Additionally, Infi-WebMath4+ outperforms InfiMM-WebMath, but plateaus after 80B tokens (roughly 10 epochs), likely due to data repetition, a trend not seen in FineMath4+.

\vspace{-0.1cm} %
\subsection{Code data}
\vspace{-0.15cm} %
\label{sec:code-datasets}
Code generation and understanding are becoming essential capabilities for modern LLMs, enabling diverse use cases such as code completion, debugging, and software design.
While specialized code models~\cite{lozhkov2024starcoder,bai2023qwen,roziere2023code} are optimized specifically for these tasks, general-purpose LLMs are increasingly deployed as coding assistants. Moreover, recent research has shown that including code data in pretraining enhances not only code-related capabilities but also improves natural language reasoning and world knowledge~\cite{aryabumi2024code}.
The Stack datasets are state-of-the-art open code datasets~\cite{li2023starcoder,kocetkov2022stack}, including Stack v1, \char`~3TB of source code from public GitHub repositories; StarCoderData~\cite{li2023starcoder,kocetkov2022stack,lozhkov2024starcoder}, a filtered subset of 250 billion tokens across 80 programming languages; Stack v2, with \char`~32TB of data sourced from the \href{https://www.softwareheritage.org/}{Software Heritage code archive}; and StarCoder2Data, the training corpus for StarCoder2 models~\cite{lozhkov2024starcoder} with 900 billion tokens spanning more than 600 programming languages.

\paragraph{Stack-Edu}
\label{sec:stack-edu}
Recent work has shown that the FineWeb-Edu classifier-based filtering strategy can be effective for code data~\cite{wei2024arctic,allal2024SmolLM}.
We therefore constructed Stack-Edu, a filtered variant of StarCoder2Data focusing on educational and well-documented code. 
Specifically, we selected the 15 largest programming languages from StarCoder2Data to match the capacity constraints of smaller models~\cite{lozhkov2024starcoder} and ensure benchmark coverage for the ablations.
This subset had \char`\~450 billion tokens.
We then trained 15 language-specific classifiers using the StarEncoder model~\cite{li2023starcoder} on synthetic annotations generated by Llama3-70B-Instruct~\cite{dubey2024llama} (prompt in~\cref{app:stack-edu-prompt}), which rated the educational quality on a scale from 0 to 5.
Each classifier was trained on 500,000 samples and achieved an F1 score above 0.7 for most languages when applying a threshold of 3 for binary classification.

To evaluate Stack-Edu, we performed annealing ablations
as described in~\cref{sec:ablation_setup}.
Filtering with a threshold of 3 improved performance across most languages while maintaining sufficient data, although Java performed better with threshold 2.
Since Markdown is not included in the MultiPL-E benchmark, we could not determine a threshold for the dataset quantitatively; instead, we used threshold 3 based on qualitative analysis.
The resulting Stack-Edu dataset contains \char`\~125B tokens across its 15 languages (see~\cref{app:stack-edu-stats}).
\cref{tab:stack-edu} shows the statistics of the top 4 programming languages in terms of size, and the impact of our educational filtering on MultiPL-E.

\begin{table}[t]
\caption{Stack-Edu dataset statistics and MultiPL-E scores for the top 4 (in terms of size) programming languages. We use HumanEval for Python evaluation.}
\vspace{-0.5pt} %
\label{tab:stack-edu}
\centering
\small
\resizebox{\columnwidth}{!}{
\begin{tabular}{@{}lrrc@{}}
\toprule
Language & StarCoder2Data & Stack-Edu & MultiPL-E \\
& (B tokens) & (B tokens) & (Original → Filtered) \\
\midrule
Python & 50.6 & 21.8 & 20.7 → 25.6 \\
C++ & 69.7 & 16.0 & 16.7 → 24.8  \\
JavaScript & 45.3 & 11.1 & 18.2 → 22.4 \\
Java & 45.6 & 42.1 & 17.6 → 22.7 \\
\bottomrule
\end{tabular}
}
\vspace{-10pt} %
\end{table}

\section{Pretraining}

\label{sec:pretraining-phases}
Recent trends in language models pretraining show a clear shift towards significantly longer training durations, especially for smaller models~\cite{Qwen2TR,Yang2024Qwen2TR,llama3_2modelcard}. While this strategy deviates from the Chinchilla-optimal guidelines~\cite{hoffmann2022training}, the resulting performance gains and reduced inference costs make extended training a worthwhile trade-off~\cite{harms_law}. For example, Qwen2-1.5B was trained on 7 trillion tokens, Qwen2.5-1.5B on 18 trillion tokens, and Llama3.2-1B, derived from a pruned 8B model, was trained using distillation on 9 trillion tokens~\cite{Qwen2TR,Yang2024Qwen2TR,llama3_2modelcard}.

When building SmolLM2, we trained on 11 trillion tokens (approximately two epochs on our collected datasets), employing a multi-stage training approach instead of a fixed dataset mixture throughout pretraining. This design was guided by four key principles: (1) \textbf{Performance-driven interventions}, where we monitor evaluation metrics on key benchmarks and adapt dataset mixtures to address specific capability bottlenecks; (2) \textbf{Upsampling high-quality math and code during the annealing phase}, reserving datasets like FineMath and parts of Stack-Edu for the final stages to maximize their impact~\cite{blakeney2024does,olmo2_2024}; (3) \textbf{Strategic introduction of medium-sized datasets}, such as OWM, InfiMM-WebMath, and Stack-Edu, mid-training to avoid dilution by larger datasets early on; and (4) \textbf{Avoiding excessive data repetition}, in line with~\citet{muennighoff2023scaling} we aimed to stay close to the recommended 4–5 epoch threshold for most datasets.
While it might be fruitful to perform multiple from-scratch training runs to explore different data mixing schedules, the high cost of pretraining SmolLM2 (around \$250,000 USD of GPU compute) motivated our ``online'' approach.

\begin{figure}
    \centering
    \includegraphics[width=1\linewidth]{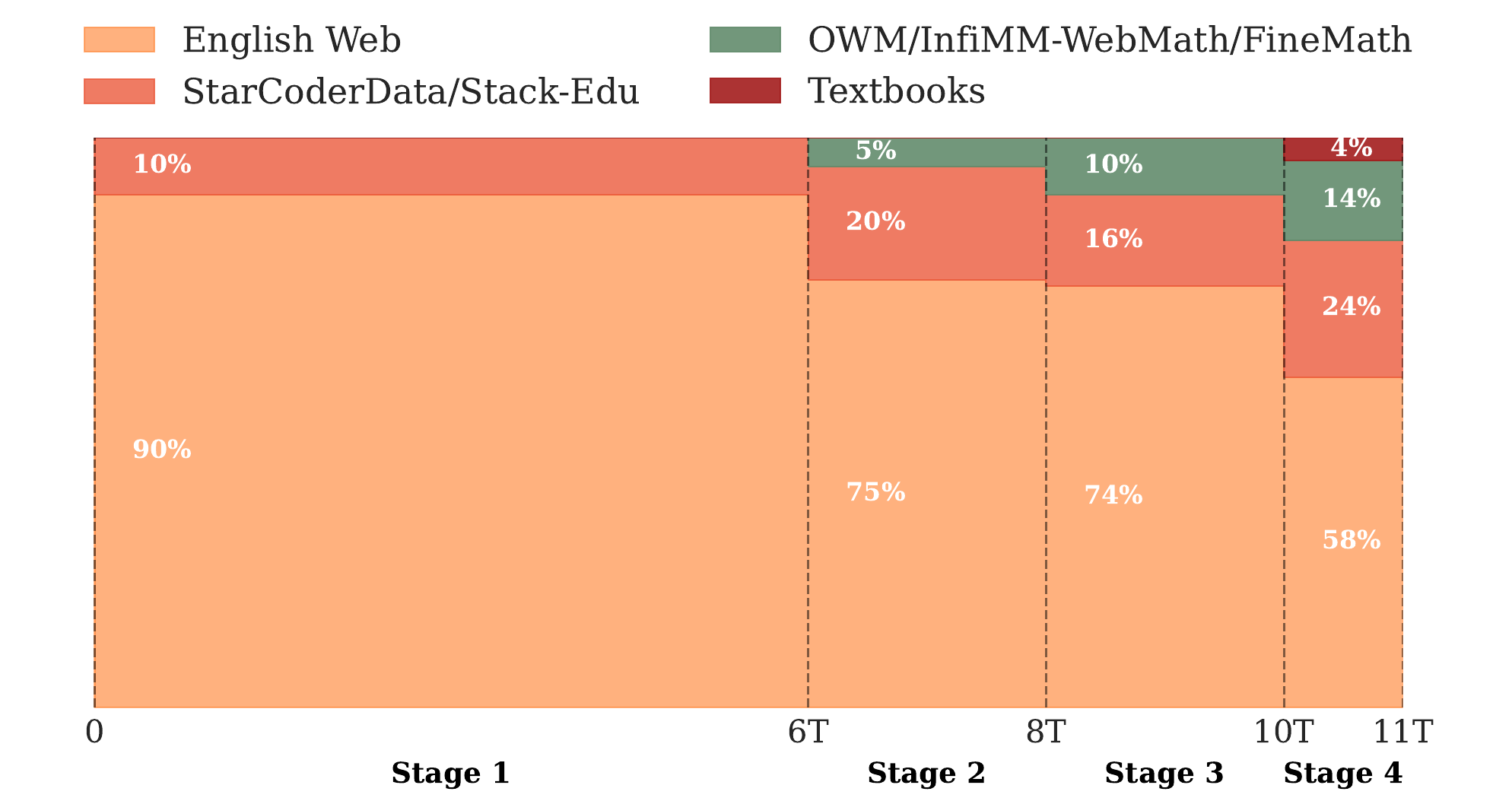}
    \caption{Dataset mixtures across training stages. Detailed descriptions are provided in~\cref{sec:pretraining-phases}. The x-axis represents the number of training tokens.}
    \label{fig:data-mixtures}
    \vspace{-1em}
\end{figure}

In the following sections, we describe each stage of the training process, detailing the dataset mixtures, the rationale behind our choices, and the observations that guided our interventions. While some decisions were informed by established findings in the literature, others were driven by empirical insights gathered during training. The data mixtures of the four pretraining phases are available in~\cref{fig:data-mixtures}.

\subsection{Training setup}
\label{subsec:training-setup}
Our base model contains 1.7B parameters and follows the LLama2 \cite{touvron2023llama} architecture, outlined in~\cref{app:training}. We trained the model on 256 H100s using the \href{https://github.com/huggingface/nanotron/}{\texttt{nanotron}} framework and use AdamW optimizer with $(\beta, \beta_2) = (0.9,0.95)$ with a Warmup Stable Decay (WSD) \cite{hu2024minicpmunveilingpotentialsmall,zhai2022scalingvisiontransformers} learning rate schedule to avoid setting a fixed training duration (see \cref{fig:wsd-scheduler},~\cref{app:training}). The schedule started with a 2,000-step warmup phase, maintained a peak learning rate of $5.0 \times 10^{-4}$ (stable phase), and could transition to a decay phase when needed, reducing the learning rate to zero over 10\% of the total training steps~\cite{hagele2024scaling}.
We used the tokenizer from \citet{allal2024SmolLM},
which has a vocabulary size of 49,152 tokens and was trained on a mixture of 70\% of FineWeb-edu, 15\% Cosmopedia-v2, 8\% OpenWebMath, 5\% StarCoderData and 2\% StackOverflow.

\subsection{Stable phase: stage 1}
\label{sec:phase1}
\paragraph{Data mixture} In the first phase of SmolLM2's pretraining (0 to 6T tokens), we designed our dataset mixture based on insights from our English web ablations and existing literature. We adopted a 60\% FineWeb-Edu and 40\% DCLM ratio (discussed in Section 2.2) for web data, which provided an optimal balance between educational content and diverse, real-world Q\&A-style data. For code data, following~\citet{aryabumi2024code}, we incorporated StarCoderData, 
consisting of 250B tokens across 80 programming languages, and limited it to 10\% of the total mixture to ensure approximately 4 epochs over 11T tokens with room for upsampling in later stages. 
We did not include math data in stage 1 due to our math datasets' relatively small size.
\vspace{-0.1cm} %
\paragraph{Findings} After 6T tokens of training, we evaluated SmolLM2 on key benchmarks, as shown in~\cref{tab:training_stages_evals_short}. Knowledge and reasoning performance aligned with expectations based on our English web ablation results. However, we observed generally poor coding and mathematics performance.%

\subsection{Stable phase: stage 2}

\vspace{-0.1cm} %
\paragraph{Data mixture} For stage 2 (6T to 8T tokens), we added OWM to the mixture at a 5\% ratio and increased the proportion of code data in hopes of maintaining strong knowledge retention while addressing observed gaps in coding and mathematical reasoning. Including OWM at a low percentage reflects the dataset's small size (12B tokens) and our gradual approach to incorporating math content. The final mixture for stage 2 consisted of 75\% English web data (keeping the 60/40 FineWeb-Edu to DCLM ratio from stage 1), 20\% code data, and 5\% math data, as shown in~\cref{fig:data-mixtures}.

\vspace{-0.25cm} %
\paragraph{Findings} After stage 2, code performance improved across most languages, validating the decision to upsample StarCoderData. OWM integration had no significant impact on math performance, underscoring the need for larger, higher-quality math datasets in later stages.
Beyond code and math performance, as shown in~\cref{fig:mmlu-progression} (\cref{app:mmlu_eval_curves}), we observed above-random  (>25\%) MMLU accuracy with a multiple-choice formulation (MCF, i.e.\ explicitly outputting an option from 'A', 'B', 'C', or 'D' instead of computing the likelihood of different answers as in the cloze formulation).
This contrasts prior work showing that small models struggle with the MCF~\cite{gu2024olmes,du2024understanding} and suggests that long trainings of small models can make them acquire abilities typically associated with larger models~\cite{blakeney2024does,gu2024olmes,du2024understanding}.
To further optimize MMLU performance, we revisited our English dataset mixture with additional annealing ablations and found that increasing DCLM relative to FineWeb-Edu slightly improves MMLU MCF at this stage.

\begin{table}[t]
\caption{Average model performance on different benchmark categories \textbf{after each training stage}. Stages 1-3 are during stable phase (no decay). Full per-benchmark results in \cref{app:evals_stages_full}.}
\label{tab:training_stages_evals_short}
\vspace{0.5em}
\resizebox{\columnwidth}{!}{
\begin{tabular}{l*{4}{c}}
\toprule
& \textbf{Stage 1} & \textbf{Stage 2} & \textbf{Stage 3} & \textbf{Stage 4} \\
Tokens & 0-6T & 6-8T & 8-10T & 10-11T \\
\midrule
Knowledge/Reasoning & 55.50 & 56.76 & 57.47 & \textbf{60.24} \\
Math & 3.21 & 3.7 & 7.27 & \textbf{22.07} \\
Code & 8.87 & 10.56 & 16.75 & \textbf{23.21} \\
Generative Tasks & 31.54 & 31.30 & 34.70 & \textbf{36.12} \\
\bottomrule
\end{tabular}
}
\vspace{-0.35cm} 

\end{table}

\subsection{Stable phase: stage 3}
\paragraph{Data mixture}

In the third and last stage of the stable phase (8T to 10T tokens, before annealing starts), we added the text-only English portion of InfiMM-WebMath with OWM, bringing the total proportion of math data to approximately 10\%, as shown in~\cref{fig:data-mixtures}. For English web data, we revisited our ablation findings and adjusted the FineWeb-Edu to DCLM ratio to 40/60. For code, we replaced StarCoderData with Stack-Edu (\cref{sec:stack-edu}). For languages with fewer than 4B tokens in Stack-Edu (TypeScript, Shell, Swift, Go, Rust and Ruby), we used their StarCoder2Data subsets. We also added Jupyter Notebooks from StarCoder2~\cite{lozhkov2024starcoder}, which provides rich, contextual examples of code interleaved with explanations, enhancing the model's reasoning around programming tasks.

\paragraph{Findings} While the integration of these new datasets brought improvements across multiple benchmarks, we observed a noticeable loss spike during this phase which remained even after rewinding training and skipping data associated with the spike~\citep{chowdhery2023palm,almazrouei2023falcon}. The exact cause remains undetermined but most evaluation metrics recovered by the end of the stage.

\vspace{-0.15cm} %
\subsection{Decay phase: stage 4}
\vspace{-0.1cm} %
\paragraph{Data mixture} The final stage consisted of decaying the learning rate linearly to 0 for 10\% of the total training duration (from 10T to 11T tokens)~\cite{hagele2024scaling}. Following~\citet{blakeney2024does}, we introduced our highest quality mathematical datasets, InfiWebMath-3+, and FineMath 4+. 
We additionally allocated 0.08\% of the mixture to OWM and 0.02\% to AugGSM8K~\cite{li2024mugglemath}, an augmented version of the GSM8K benchmark's training set, which has become a common component of recent pretraining datasets~\cite{achiam2023gpt,dubey2024llama,olmo2_2024}.
Overall, mathematical content totaled 14\% of the mixture.
We expanded Stack-Edu to include additional programming languages not covered in stage 3, and set the dataset's contribution to 24\% of the mixture. We maintained the natural distribution across programming languages, with a higher allocation for Python. The remaining mixture consisted of English web data at 58\% (maintaining the higher DCLM to FineWeb-Edu ratio) and Cosmopedia v2~\cite{allal2024SmolLM} at 4\%, which provides 30B tokens of high-quality synthetic textbooks, blog posts, and stories.

\vspace{-0.2cm} %
\paragraph{Findings} While all benchmark tasks show improvements after stage 4, we observe substantial gains in coding performance and, most notably, in math performance, validating our data mixture specifically targeting these domains.

\vspace{-0.1cm} %
\subsection{Context Length extension}
\vspace{-0.1cm} %

To support long-context applications, we followed standard practice~\cite{gao2024trainlongcontextlanguagemodels} and extended the context length from 2k to 8k tokens, by taking an intermediate checkpoint from stage 4 (before the final 75 billion tokens of training) and continuing training with a different data mixture and a RoPE value of 130k.
The mixture was adjusted to include 40\% long-context documents (8k tokens or more) sourced from DCLM (10\%), FineWeb-Edu (10\%), and the books subset of Dolma (20\%)~\cite{soldaini2024dolma}, while the remaining 60\% followed the stage 4 mixture. After this step, we obtain the final SmolLM2 base model.

\vspace{-0.25cm} %
\subsection{Base model evaluation}
\vspace{-0.1cm} %
\label{sec:evaluation-basemodels}
We evaluate and compare the final base SmolLM2 model with existing state-of-the-art models of similar size, Qwen2.5-1.5B~\cite{Yang2024Qwen2TR} and Llama3.2-1B~\cite{llama3.2}, on a wide range of benchmarks. Evaluations are conducted using \href{https://github.com/huggingface/lighteval/}{\texttt{lighteval}} and in a zero-shot setting unless otherwise specified.

\vspace{-0.05cm} %
Evaluation results in~\cref{tab:base_model_evals} show the strong performance of base SmolLM2, outperforming the Qwen2.5 base model on HellaSwag, and ARC. SmolLM2 also delivers strong performance on held-out benchmarks not monitored during training, such as MMLU-Pro~\citep{wang2024mmlu},  TriviaQA~\cite{joshi2017triviaqa}, and Natural Questions (NQ, \citealp{kwiatkowski2019natural}). Notably, the model outperforms Qwen2.5-1.5B by nearly 6 percentage points on MMLU-Pro, further validating its generalization capabilities.
On math and coding benchmarks, SmolLM2 demonstrates competitive performance. While it lags behind Qwen2.5-1.5B, SmolLM2 outperforms Llama3.2-1B on GSM8K, MATH and HumanEval. 
Importantly, we see next to no degradation in performance after Context Length Extension, while the HELMET~\cite{yen2024helmetevaluatelongcontextlanguage} and 
Needle in the Haystack (NIAH)~\cite{kamradt2024needle} results show strong performance -- see~\cref{app:long-context-evaluations}. These results highlight the effectiveness of our curated datasets, data mixtures, and training stages. 
\begin{table}[t]
\caption{Performance comparison of SmolLM2 and other 1-2B \textbf{base models} across benchmarks. SmolLM2 demonstrates competitive results  highlighting its generalization capabilities.}
\label{tab:base_model_evals}
\vspace{0.5em}
\resizebox{\columnwidth}{!}{
\begin{tabular}{l*{3}{c}}
\toprule
Model family & \textbf{SmolLM2} & \textbf{Llama3.2} & \textbf{Qwen2.5} \\
Parameters & 1.7B & 1B & 1.5B \\
\midrule
HellaSwag        & \textbf{68.7} & 61.2 & 66.4 \\
ARC     & \textbf{60.5} & 49.2 & 58.5 \\
PIQA             & \textbf{77.6} & 74.8 & 76.1 \\
CommonsenseQA    & \textbf{43.6} & 41.2 & 34.1 \\
Winogrande       & \textbf{59.4} & 57.8 & 59.3 \\
OpenBookQA       & \textbf{42.2}         & 38.4 & 40.0 \\
\midrule
MMLU-Pro (held-out)  & \textbf{19.4} & 11.7 & 13.7 \\
Natural Questions (held-out)       & 8.7 & 6.2 & \textbf{10.5}\\
TriviaQA  (held-out)       & \textbf{36.7} & 28.1 & 20.9 \\
\midrule
GSM8K (5-shot)   & 31.1          & 7.6  & \textbf{61.7} \\
MATH (4-shot)   & 11.6          & 3.3  & \textbf{34.3} \\ %
HumanEval    & 22.6          & 18.9  & \textbf{37.2} \\
\bottomrule
\end{tabular}
}
\vspace{-0.65cm} %
\end{table}
\vspace{-0.25cm} %

\section{Post-training}
\vspace{-0.15cm} %

\label{sec:instruction-tuning}

After training the base SmolLM2 model, we followed current standard practice for maximizing performance and utility via post-training through instruction tuning and preference learning. 
For post-training, we leveraged existing datasets in addition to a new instruction tuning dataset called SmolTalk.

\vspace{-0.1cm} %
\subsection{SmolTalk}
\vspace{-0.15cm} %
Although the SmolLM2 base model outperformed other state-of-the-art base models in the 1-2B parameter range, the base model's performance after fine-tuning on public datasets like MagPie-Pro~\cite{xu2024magpie} or OpenHermes2.5~\cite{OpenHermes2.5} was lower than the post-trained versions of these other models.
This observation motivated the development of SmolTalk\footnote{\href{https://huggingface.co/datasets/HuggingFaceTB/smoltalk}{https://huggingface.co/datasets/HuggingFaceTB/smoltalk}}, a new instruction-following dataset that carefully combines selected existing datasets with new synthetic datasets we developed, including the Magpie-Ultra conversational dataset as well as other specialized datasets that address specific capabilities like Smol-Constraint, Smol-Rewrite, and Smol-Summarization. All datasets were generated using Distilabel~\cite{distilabel-argilla-2024}.

\subsubsection{Conversational data}
MagPie-Ultra is a multi-turn dataset created using the two-step prompting method from~\cite{xu2024magpie}. Unlike MagPie, which used Llama-3-70B-Instruct without specific system prompts to generate two-turn conversations, MagPie-Ultra leverages the larger, more powerful model Llama-3.1-405B-Instruct-FP8~\cite{dubey2024llama}. We also incorporate system prompts to guide generation, producing a balanced dataset of 1M samples with three-turn conversations.
The resulting dataset was further filtered using smaller Llama models (Llama-3.1-8B-Instruct and Llama-Guard-3-8B) to ensure quality and safety of the generated instructions. We also leveraged ArmoRM \cite{ArmoRM,wang2024arithmetic} to score conversations for quality-based filtering, and gte-large-en-v1.5 \cite{zhang2024mgte,li2023towards} to deduplicate semantically similar conversations.

We compare MagPie-Ultra to existing public supervised fine-tuning (SFT) datasets in~\cref{tab:instruction-tuning-ablations} (\cref{app:post-training}). The evaluation suite included the instruction-following and conversation benchmarks IFEval~\cite{zhou2023instruction} and MT-Bench~\cite{zheng2023judging}; reasoning in ARC Challenge; knowledge in MMLU-Pro as well as GSM8K and MATH for math evaluations. Our dataset outperforms MagPie-Pro on most benchmarks, and largely surpasses OpenHermes2.5 and UltraChat~\cite{ding2023enhancing} on IFEval and MT-Bench.

\subsubsection{Task-specific data}
We developed additional task-specific datasets to further enhance model instruction-following with detailed constraints (Smol-Constraint), summarization (Smol-Summarization) and rewriting (Smol-Rewrite) capabilities.
Smol-Constraint contains 36k instructions with detailed constraints similar to the ones found in IFEval \cite{zhou2023instruction}. Using the method from~\cite{xu2024magpie} with a targeted system prompt, we generated 550k instructions and responses for these instructions using Qwen2.5-72B-Instruct~\cite{Yang2024Qwen2TR}. We then filtered out generated instructions that contained conflicting constraints or incorrect responses, resulting in 56.3k instruction-response pairs, which after decontaminating against IFEval (10 n-gram overlap), yielded 36k pairs.
For Smol-Summarization and Smol-Rewrite, we first generated high-quality source texts that would serve as the basis for summarization and rewriting tasks. We synthesized a diverse collection of emails, tweets, LinkedIn posts, and notes using PersonaHub~\cite{ge2024scaling} and personas from the FinePersonas dataset~\cite{finepersonas2024,chan2024scalingsyntheticdatacreation}. This allowed us to generate diverse content by prompting Qwen2.5-72B-Instruct with specific system prompts and a persona description, obtaining texts with various writing styles, topics and perspectives. We then prompted Qwen2.5-72B-Instruct to summarize and rewrite the given texts, obtaining around 1M summaries and 600k rewritten texts.
Adding the 3 Smol- datasets to MagPie-Ultra, (MagPie-Ultra$\overset{+}{}$) further improves IFEval performance as shown in ~\cref{tab:instruction-tuning-ablations} (\cref{app:post-training}).

\vspace{-0.1cm} %
\subsubsection{Math data}
\vspace{-0.15cm} %
To improve mathematical reasoning, we evaluated public math instruction datasets by fine-tuning on mixtures with 80\% general instruction data (MagPie Ultra + Smol-Constraint, Smol-Rewrite, Smol-Summarization) and 20\% math data from various sources. Results in~\cref{tab:instruction-tuning-ablations} (\cref{app:post-training}) highlight complementary dataset strengths: NuminaMath-CoT~\cite{numina_math_datasets} demonstrated strong performance on MATH and MT-Bench, while MetaMathQA~\cite{yu2023metamath}, which is also included in OpenHermes2.5, improved results on GSM8K. Based on these findings, we incorporated a combination of both datasets into SmolTalk.
\vspace{-0.1cm} %
\subsubsection{Other specialized data}
\vspace{-0.15cm} %
For code generation, we used Self-OSS-Starcoder2-Instruct~\cite{wei2024selfcodealign}, containing 50k high-quality Python instruction-response pairs. To support system prompts, we included 30k randomly selected samples from SystemChats2.0~\cite{systemchat2024}, and for function calling, we added 80k samples from APIGen-Function-Calling~\cite{liu2024apigen}. Additionally, to maintain strong performance on long-context tasks, we incorporated an English subset of LongAlign~\cite{bai2024longalign} (3.7k samples with 8k–16k tokens). We also added 100k randomly selected OpenHermes2.5 samples due to its strong performance in knowledge (MMLU-Pro), Everyday-Conversations~\cite{everydayconversations2024}, 2.2k casual multi-turn interactions, and Explore-Instruct~\cite{wan2023explore} for rewriting. We found that incorporating these datasets with the specified number of samples effectively enhanced their target capabilities while preserving strong performance across other benchmarks.
\vspace{-0.1cm} %
\subsection{Supervised fine-tuning (SFT)}
\vspace{-0.1cm} %
\cref{tab:smoltalk_composition} (\cref{app:post-training}) shows the final composition of SmolTalk. We performed supervised fine-tuning of our base SmolLM2 on SmolTalk for 2 epochs, using a global batch size of 128, sequence length of 8192, and a learning rate of $3.0 \times 10^{-4}$. The evaluation results after this SFT phase are available in~\cref{tab:instruction-tuning-ablations} (\cref{app:post-training}).

\vspace{-0.1cm} %
\subsection{Alignment}
\vspace{-0.1cm} %

For preference learning, we used Direct Preference Optimization (DPO)~\cite{rafailov2024direct}. We experimented with various public synthetic feedback datasets~\cite{ivison2024unpacking} including UltraFeedback~\cite{cui2024ultrafeedback}, UltraInteract~\cite{yuan2024advancing}, Capybara~\cite{daniele2023amplify-instruct}, and ORCA~\cite{lv2023orca}.
UltraFeedback proved the most consistently effective across benchmarks, improving MT-Bench, MMLU-Pro, and MATH. We trained for 2 epochs with a learning rate of $1.0 \times 10^{-6}$, beta of 0.5, global batch size of 128, and sequence length of 1024 tokens. After this final stage of DPO training, we obtain the instruct SmolLM2 model. As noted in~\citet{dubey2024llama}, using short-context data for DPO did not impact the model’s 8k context ability.

\vspace{-0.1cm}
\subsection{Instruct model evaluation}
\vspace{-0.1cm}
\label{sec:evaluation_smollm2}
We evaluate the final instruct version of SmolLM2 and compare it with the instruct variants of Qwen2.5-1.5B and Llama3.2-1B, with results shown in~\cref{tab:instruct_model_evals}. 
SmolLM2-Instruct shows strong instruction following capabilities, strongly outperforming Qwen2.5-1.5B-Instruct on IFEval; our model is competitive on MT-Bench and OpenRewrite-Eval~\cite{shu2024rewritelm} for text rewriting, and demonstrates strong mathematical capabilities as evidenced by the GSM8K and MATH scores. These results highlight SmolLM2's ability to generalize across a variety of tasks, showcasing its potential as a capable chat assistant.

\begin{table}[t]
\caption{Comparison of 1-2B \textbf{instruction-tuned models} across benchmarks. SmolLM2-1.7B-Instruct exhibits strong performance in instruction-following, reasoning, and math.}
\label{tab:instruct_model_evals}
\vspace{0.5em}
\resizebox{\columnwidth}{!}{
\begin{tabular}{l*{3}{c}}
\toprule
\textbf{Model} & \textbf{SmolLM2-1.7B} & \textbf{Llama3.2-1B} & \textbf{Qwen2.5-1.5B} \\
\midrule
IFEval (Average)    & \textbf{56.7} & 53.5 & 47.4 \\
MT-Bench         & 6.13          & 5.48 & \textbf{6.52} \\
OpenRewrite-Eval & 44.9          & 39.2 & \textbf{46.9} \\
ARC    & \textbf{51.7} & 41.6 & 46.2 \\
BBH (3-shot)     & 32.2          & 27.6 & \textbf{35.3} \\
MMLU-Pro    & 19.3          & 12.7 & \textbf{24.2} \\
HellaSwag        & \textbf{66.1} & 56.1 & 60.9 \\
PIQA             & \textbf{74.4} & 72.3 & 73.2 \\
\midrule
GSM8K (5-shot)   & 48.8 & 37.4 & \textbf{63.3} \\
MATH (4-shot)   & \textbf{21.0} & 19.5 & 19.6 \\ %
HumanEval   & 28.1 & \textbf{33.5} & 30.5 \\
\bottomrule
\end{tabular}
}
\vspace{-1em}
\end{table}

\section{SmolLM2 135M and 360M}
In addition to SmolLM2-1.7B, we also trained two smaller models: SmolLM2-360M (360M parameters, trained on 4T tokens) and SmolLM2-135M (135M parameters, trained on 2T tokens), which are similarly state-of-the-art in their size class. Given their smaller capacity and reduced training cost, we re-ran data ablations at the target training length to determine the most effective data mixture. We found that filtering DCLM with the FineWeb-Edu classifier, removing samples with score 0, and downsampling those with scores 1 and 2 worked best. Unlike SmolLM2-1.7B, where we leveraged a multi-stage training strategy, these smaller models benefited from a single-stage training approach with consistently high-quality data. We incorporated Stack-Edu from the start, alongside InfiMM-WebMath, FineMath, and Cosmopedia. These models share the same architecture as SmolLM2-1.7B but use Grouped Query Attention (GQA) and were trained using the WSD scheduler with 20\% decay and a learning rate of $3.0 \times 10^{-3}$. For post-training, we applied SFT using a filtered version of SmolTalk\footnote{\href{https://huggingface.co/datasets/HuggingFaceTB/smol-smoltalk}{https://huggingface.co/datasets/HuggingFaceTB/smol-smoltalk}}, removing complex instruction-following tasks (e.g., function calling) and hard examples from MagPie-Ultra to better align with the models' capacity. Finally, we performed DPO training using UltraFeedback, optimizing the models for instruction-following while preserving coherence and helpfulness. More details about SmolLM2-360M and 135M can be found in their respective model cards\footnote{\href{https://huggingface.co/HuggingFaceTB/SmolLM2-360M-Instruct}{SmolLM2-360M model card}}\footnote{\href{https://huggingface.co/HuggingFaceTB/SmolLM2-135M-Instruct}{SmolLM2-135M model card}}.

\vspace{-0.05cm} %
\section{Conclusion}
\vspace{-0.05cm} %

SmolLM2 advances the state-of-the-art for open small LMs through a combination of careful dataset curation and multi-stage training. Our approach highlights the critical role of high-quality, specialized datasets in enabling smaller models to achieve strong performance across a variety of benchmarks. The development of FineMath, Stack-Edu, and SmolTalk addressed limitations in existing public datasets, improving capabilities in reasoning, mathematics, and instruction-following tasks. To support future research and development, we release SmolLM2 alongside the datasets and code used in its training. These resources provide a comprehensive foundation for training performant small language models, making them accessible to a broader range of researchers and applications.

\section*{Acknowledgments}
\label{app:acknowledgments}

This work would not have been possible without the contributions and support of our collaborators and colleagues:

\begin{itemize}
    \item We thank Nouamane Tazi, Phuc Nguyen, Ferdinand Mom, and Haojun Zhao for designing and building our training framework, Nanotron.
    \item We thank Guilherme Penedo and Hynek Kydlíček for building our data pipeline framework, Datatrove.
    \item We thank Clémentine Fourrier and Nathan Habib for developing our evaluation framework, LightEval.
    \item We thank all our colleagues who participated in discussions that contributed to the development and refinement of SmolLM2.
    \item We thank Muhammed Emin Baslak and Pierre-Carl Langlais for their work on enhancing SmolLM2 with Entropix.
\end{itemize}

We also extend our gratitude to the broader research community and open-source ecosystem for fostering collaboration and innovation, without which this project would not have been possible.

\section*{Impact Statement}
This paper presents work whose goal is to advance the field of Machine Learning. There are many potential societal consequences of our work, none which we feel must be specifically highlighted here.

\bibliography{references}
\bibliographystyle{icml2025}

\newpage
\appendix
\onecolumn

\section{Training setup}
\label{app:training}
\cref{tab:model_config} shows the architecture details of SmolLM2 1.7B.
\begin{table}[h!]
\centering
\caption{Overview of the architecture of SmolLM2. $^\dagger$ This is before extending the context to 8k tokens.}
\label{tab:model_config}
\begin{tabular}{@{}lcc@{}}
\toprule
\textbf{Parameter}          & \textbf{Value} \\ \midrule
Layers                      & 24            \\
Model Dimension             & 2,048         \\
FFN Dimension               & 8,192         \\
Attention Heads             & 32            \\
Sequence Length             & 2,048 $^\dagger$        \\
Token per batch        & 2M \\
Tied embedding              & Yes         \\                  
Positional Embeddings       & RoPE ($\theta = 10,000$) \\
Activation Function         & SwiGLU          \\ \bottomrule
\end{tabular}
\end{table}

\begin{figure}[h]
    \centering
    \includegraphics[width=0.6\linewidth]{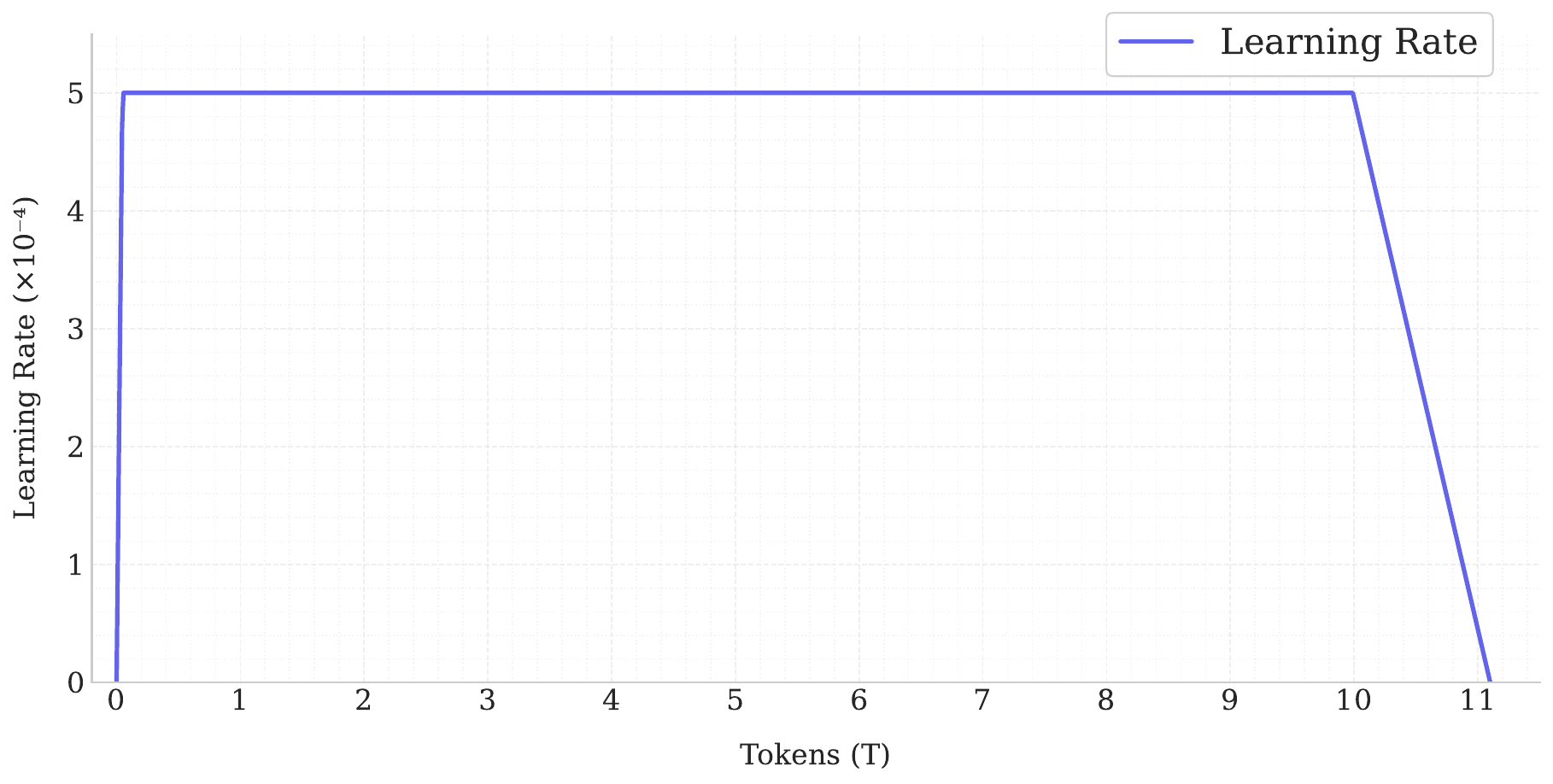}
    \caption{Learning rate during SmolLM2 training. We used WSD scheduler with 2000 steps warmup, learning rate $5.0 \times 10^{-4}$ and 10\% decay.}
    \label{fig:wsd-scheduler}
\end{figure}

\cref{fig:wsd-scheduler} shows the progression of the learning rate through the training using WSD scheduler.

\clearpage
\section{English web ablations}
\label{app:english-web}
\begin{figure}[h]
    \centering
    \includegraphics[width=0.8\linewidth]{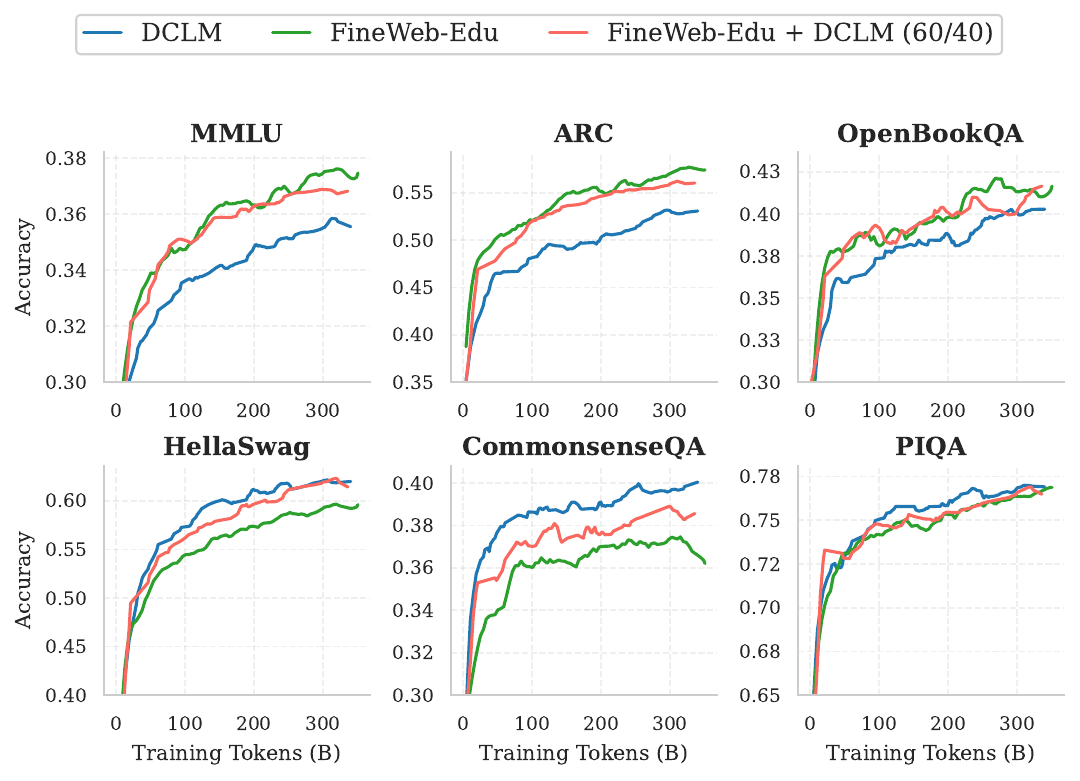}
    \caption{Evaluation of models trained on FineWeb-Edu and DCLM for 350B tokens. FineWeb-Edu excels at knowledge and reasoning tasks, while DCLM demonstrates stronger performance on commonsense reasoning benchmarks. A 60/40 mixture of FineWeb-Edu and DCLM achieves balanced performance across all tasks.}
    \label{fig:fwedu-vs-dclm-curves}
\end{figure}

\cref{fig:fwedu-vs-dclm-curves} shows the evaluation curves of ablation models trained on DCLM, FineWeb-Edu and their mix for 350B tokens.

\clearpage
\section{FineMath}

\subsection{Public datasets comparison}\label{app:owm-vs-inf}
\begin{figure}[h]
    \centering
    \includegraphics[width=0.7\linewidth]{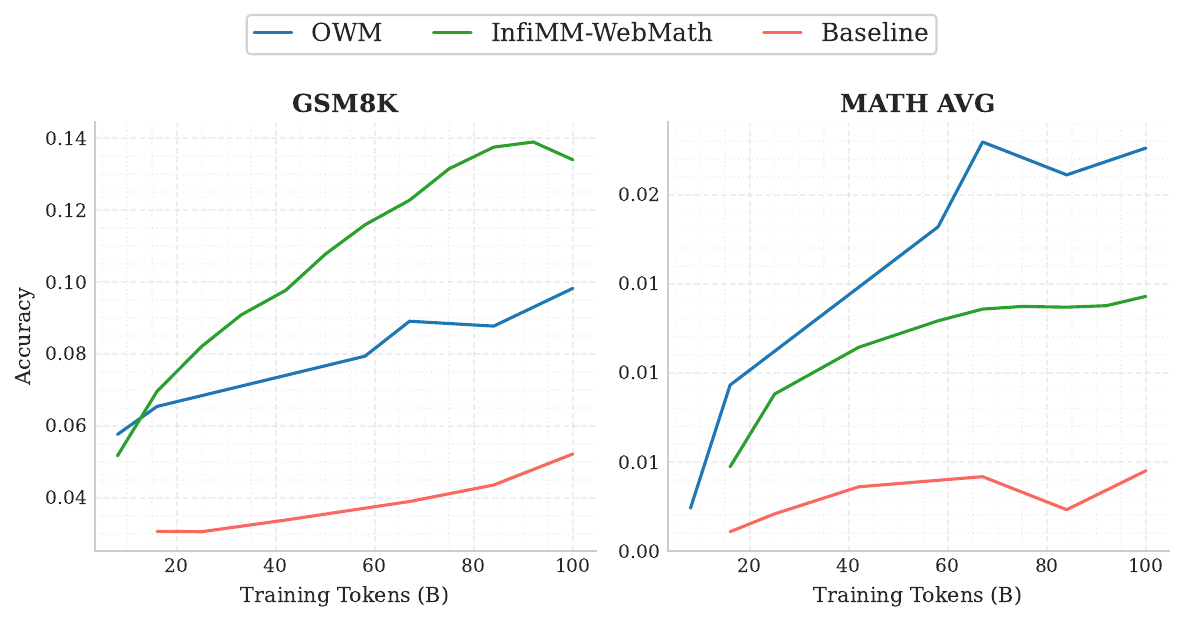}
    \caption{Results of annealing ablations comparing OWM and the text component of InfiMM-WebMath. InfiMM-WebMath consistently outperforms OWM on GSM8K, while OWM has a slight advantage on MATH. Despite training on 60B math tokens (equivalent to 5 epochs for OWM and 1.5 epochs for InfiMM-WebMath), performance remains far below state-of-the-art LLMs, highlighting the need for a new math dataset.}
    \label{fig:owm-vs-infi}
\end{figure}

\cref{fig:owm-vs-infi} Shows the performance of ablation models trained on OWM and InfiMM-WebMath on GSM8k and MATH.

\subsection{Annotation Prompt (3-scale)}\label{app:finemath-silver-prompt}

We used the following prompt template to generate the silver 3-scale annotations for FineMath using the Llama3 model:

\begin{quote}
\footnotesize
Evaluate the following text extract for its potential usefulness for studying mathematics up to high school and early undergraduate levels. Use the following 3-point scoring system described below. Points are accumulated based on the satisfaction of each criterion:

- Add 1 point if the extract contains some mathematical content, even if it's not very useful for studying or is an academic paper that is too advanced. 

- Add another point if the extract demonstrates logical reasoning in a mathematical context, even if it lacks step-by-step explanations or is too advanced.

- Award a third point if the extract is at an appropriate level (up to high school and early undergraduate levels) and contains clear mathematical deductions and step-by-step solutions to mathematical problems.

Question-answer formats (e.g., from educational websites or forums) are acceptable if they meet the criteria. Ignore any formatting errors or missing equations and make assumptions based on the overall content.

The text extract:

<EXTRACT>

After examining the extract: 

- Briefly justify your total score, up to 100 words.

- Conclude with the score using the format: "Final score:  <total points>".
\end{quote}

\subsection{Annotation Prompt (5-scale)}\label{app:finemath-gold-prompt}

We used the following prompt template to generate the 5-scale annotations for FineMath using the Llama3 model during the second filtering stage:

\begin{quote}
\footnotesize
Evaluate the following text extract for its potential usefulness for studying mathematics up to high school and early undergraduate levels. Use the following 5-point scoring system described below. Points are accumulated based on the satisfaction of each criterion:

- Add 1 point if the extract contains some mathematical content, even if it's not very useful for studying, or if it contains non-academic content such as advertisements and generated pages for converting weight and currencies.

- Add another point if the extract touches on mathematical topics, even if it's poorly written if it's too complex such as an academic paper that is too advanced. 

- Award a third point if the extract demonstrates problem solving or logical reasoning in a mathematical context, even if it lacks step-by-step explanations.

- Grant a fourth point if the extract is at an appropriate level (up to high school and early undergraduate levels) and contains clear mathematical deductions and step-by-step solutions to mathematical problems. It should be similar to a chapter from a textbook or a tutorial.

- Give a fifth point if the extract is outstanding in its educational value for teaching and studying mathematics in middle school and high school. It should include very detailed and easy to follow explanations.

Question-answer formats (e.g., from educational websites or forums) are acceptable if they meet the criteria.

The text extract:

<EXTRACT>

After examining the extract:

- Briefly justify your total score, up to 100 words.

- Conclude with the score using the format: Final score: <total points>.
\end{quote}

\clearpage
\section{Stack-Edu}
\label{app:stack-edu}

\subsection{Annotation Prompt}\label{app:stack-edu-prompt}

We used the following prompt template to generate the 5-scale annotations for Stack-Edu (Python in this case) using the Llama3 model:

\begin{quote}
\footnotesize
Below is an extract from a Python program. Evaluate whether it has a high educational value and could help teach coding. Use the additive 5-point scoring system described below. Points are accumulated based on the satisfaction of each criterion:

- Add 1 point if the program contains valid Python code, even if it's not educational, like boilerplate code, configs, and niche concepts.

- Add another point if the program addresses practical concepts, even if it lacks comments.

- Award a third point if the program is suitable for educational use and introduces key concepts in programming, even if the topic is advanced (e.g., deep learning). The code should be well-structured and contain some comments. 

- Give a fourth point if the program is self-contained and highly relevant to teaching programming. It should be similar to a school exercise, a tutorial, or a Python course section.

- Grant a fifth point if the program is outstanding in its educational value and is perfectly suited for teaching programming. It should be well-written, easy to understand, and contain step-by-step explanations and comments.

The extract:
<EXAMPLE>

After examining the extract: 

- Briefly justify your total score, up to 100 words.

- Conclude with the score using the format: "Educational score:  <total points>
\end{quote}
We use similar prompts for the other 14 programming languages in Stack-Edu, adjusting the examples in the third criterion to reflect language-specific topics. For instance, in the JavaScript prompt, we replace "deep learning" with "asynchronous programming".

\subsection{Stack-Edu language statistics}\label{app:stack-edu-stats}
\cref{tab:stack-edu-all-stats} shows the size of each programming language in Stack-Edu before and after the educational filtering. Initially, we also included HTML, but the classifier performed poorly, so we retained StarCoder2Data.
\begin{table}[h]
\caption{Stack-Edu dataset statistics across programming languages. The table shows the original dataset size (from StarCoder2Data) and filtered Stack-Edu size for each programming language.}
\label{tab:stack-edu-all-stats}
\centering
\small
\begin{tabular}{@{}lrr@{}}
\toprule
Language & StarCoder2Data & Stack-Edu \\
& (B tokens) & (B tokens) \\
\midrule
Python & 50.6 & 21.8 \\
Cpp & 69.7 & 16.0 \\
Markdown & 80.4 & 14.0 \\
C & 38.4 & 11.1 \\
JavaScript & 45.3 & 11.1 \\
Java & 45.6 & 42.1  \\
SQL & 13.7 & 9.62 \\
PHP & 44.9 & 9.07 \\
C-Sharp & 33.4 & 8.87 \\
TypeScript & 12.2 & 3.03 \\
Shell & 4.17 & 3.13 \\
Swift & 3.71 & 1.83 \\
Go & 3.67 & 1.80 \\
Rust & 3.39 & 1.75 \\
Ruby & 5.76 & 1.61 \\
\bottomrule
\end{tabular}
\end{table}

\clearpage
\section{Detailed pretraining results}
\subsection{Evaluation after each training stage}
\cref{tab:training_stages_evals} shows the evaluation results of SmolLM2 at the end of each training stage. In addition to the benchmarks used during the ablations, we added four generative tasks: CoQA~\cite{reddy2019coqa}, DROP~\cite{dua2019drop}, Jeopardy~\cite{mosaicml_jeopardy} and SQuAD v2~\cite{rajpurkar2018know}
\label{app:evals_stages_full}


\begin{table}[!h]
\caption{Per-benchmark model performance across training stages. Stages 1-3 are during stable phase (no learning rate decay).}
\label{tab:training_stages_evals}
\centering
\begin{tabular}{lcccc}
\toprule
& \textbf{Stage 1} & \textbf{Stage 2} & \textbf{Stage 3} & \textbf{Stage 4} \\
Tokens & 0-6T & 6-8T & 8-10T & 10-11T \\
\midrule
MMLU (MCF) & 29.62 & 37.96 & 42.54 & \textbf{48.87} \\
HellaSwag & 66.17 & 65.29 & 66.29 & \textbf{69.26} \\
ARC & 59.95 & 60.08 & 58.66 & \textbf{60.99} \\
OpenBookQA & 42.00 & 42.40 & 41.40 & \textbf{43.60} \\
WinoGrande & 58.88 & 58.33 & 58.64 & \textbf{61.09} \\
PIQA & 76.39 & 76.50 & 77.26 & \textbf{77.64} \\
\midrule
GSM8K & 4.32 & 4.62 & 10.01 & \textbf{32.60} \\  
MATH & 2.1 & 2.78 & 4.52 & \textbf{11.54} \\

\midrule
HumanEval & 10.97 & 9.15 & 17.68 & \textbf{22.60} \\
Multiple-E Java & 5.70 & 10.12 & 14.56 & \textbf{23.42} \\
Multiple-E JS & 9.94 & 12.42 & 18.01 & \textbf{23.60} \\
\midrule
CoQA & 33.43 & 33.98 & 38.82 & \textbf{40.45} \\
DROP & 13.69 & 11.36 & 17.19 & \textbf{19.22} \\
Jeopardy & 23.1 & 22.4 & \textbf{25.54 }& 23.35 \\
SQuAD v2 & 55.97 & 57.45 & 57.26 & \textbf{61.48} \\
\bottomrule
\end{tabular}
\end{table}

\subsection{MMLU progression}
\label{app:mmlu_eval_curves}
\begin{figure}[h]
    \centering
    \includegraphics[width=0.8\linewidth]{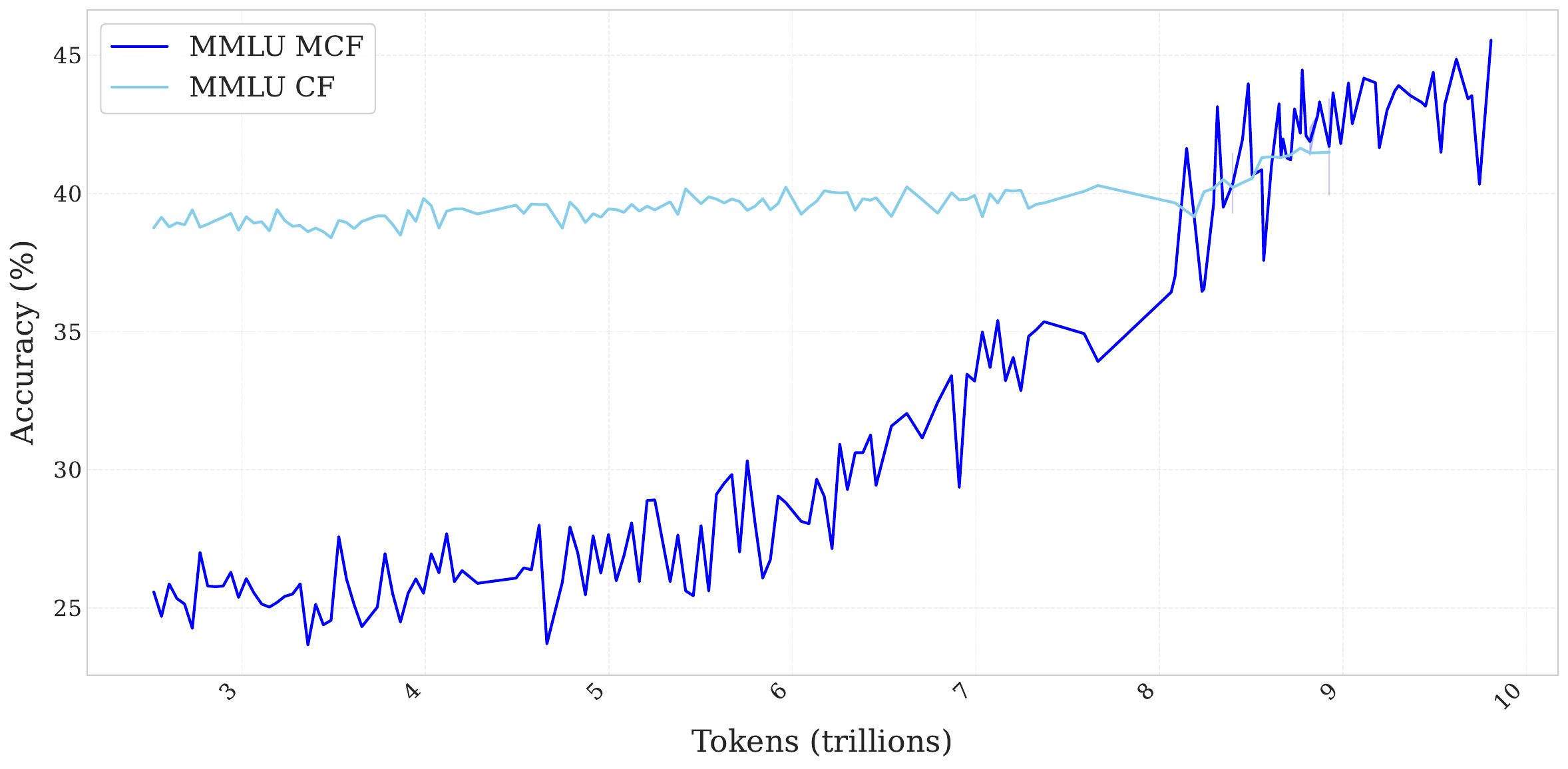}
    \caption{Progression of MMLU MCF and MLU CF during the training. We observe above-random  (>25\%) accuracy on MMLU MCF after 6T tokens of training, while MMLU CF appears to plateau.}
    \label{fig:mmlu-progression}
\end{figure}

\cref{fig:mmlu-progression} shows the progression of MMLU scores throughout the stable phase.
\clearpage

\section{Post-training}
\label{app:post-training}

\cref{tab:smoltalk_composition} shows the final composition of SmolTalk dataset.
\begin{table}[h!]
\centering
\caption{Composition of the SmolTalk dataset. The total dataset contains 1.1M instruction-response pairs from different data sources.}
\label{tab:smoltalk_composition}
\begin{tabular}{@{}lc@{}}
\toprule
\textbf{Dataset source} & \textbf{Number of samples in SmolTalk} \\ \midrule
\multicolumn{2}{c}{\textit{New datasets}} \\
\cmidrule{1-2}
MagPie-Ultra & 431k \\
Smol-Rewrite & 56.2k \\
Smol-Constraints & 36.2k \\
Smol-Summarization & 101k \\
\midrule
\multicolumn{2}{c}{\textit{Math data}}\\
\cmidrule{1-2}
NuminaMath-CoT & 112k \\
MetaMathQA & 50k \\
\midrule
\multicolumn{2}{c}{\textit{Other}} \\
\cmidrule{1-2}
Self-OSS-Starcoder2-Instruct & 50.7k \\
APIGen-Function-Calling & 87.5k \\
SystemChats2.0 & 35.9k \\
LongAlign & 3.73k \\
Everyday-Conversations & 2.38k \\
Explore-Instruct-Rewriting & 32k \\
OpenHermes2.5 & 100k \\ \midrule
\textbf{Total} & \textbf{1.1M} \\ \bottomrule
\end{tabular}
\end{table}

\cref{tab:instruction-tuning-ablations} shows the performance after training on the different components of SmolTalk we consider. The top section compares the results of fine-tuning SmolLM2 base on different instruction datasets, while the bottom section evaluates the impact of adding 20\% specialized math data to a base mixture of 80\% MagPie-Ultra$\overset{+}{}$ during the SFT. The last row, SmolLM2-SFT, represents the final SFT checkpoint of SmolLM2 before DPO, trained for two epochs on the full SmolTalk dataset.

\begin{table}[h]
\caption{Performance on instruction-tuning datasets. MagPie-Ultra$\overset{+}{}$ refers to MagPie-Ultra combined with Smol-Constraints, Smol-Rewrite, and Smol-Summarization. MagPie-Pro-MT is multi-turn while MagPie-Pro is the single turn version. 
All comparisons were performed by fine-tuning the SmolLM2 base model on each dataset for 1 epoch. SmolLM2-SFT$^\dagger$, the final supervised fine-tuned version of SmolLM2, was trained for 2 epochs on SmolTalk.
}
\label{tab:instruction-tuning-ablations}
\centering
\begin{tabular}{l r r r r r r}
\toprule
\textbf{Dataset} & \textbf{IFEval} & \textbf{MTB} 
                 & \textbf{GSM8K}  & \textbf{MATH} & \textbf{ARC-C} & \textbf{MMLU-Pro} \\
\midrule
\multicolumn{7}{c}{Instruction datasets comparison}\\
\cmidrule{1-7}
OpenHermes                    & 30.01 & 1.02 & \textbf{42.91}  & 12.76 & 40.27 & \textbf{20.32} \\
UltraChat                     & 27.26 & 4.66 & 30.40 & 9.06 & \textbf{41.21} & 15.79 \\
MagPie-Pro                    & 30.45 & 4.31 & 14.56 & 6.64 & 36.01 & 12.19  \\
MagPie-Pro-MT                    & 31.66 & \textbf{5.40} & 20.55 & 7.84 & 36.69 & 11.97 \\
MagPie-Ultra                & 35.49 & 5.22 & 24.34 & \textbf{13.56} & 37.71 & 12.01  \\
MagPie-Ultra$\overset{+}{}$                & \textbf{48.16} & 5.28 & 19.94 & 12.74 & 38.91 & 12.43 \\
\midrule  
\multicolumn{7}{c}{Math datasets comparison} \\
\cmidrule{1-7}
MagPie-Ultra$\overset{+}{}$ 
 + MathInstruct               & \textbf{47.05} & 5.43 & 30.1 & 14.0 & \textbf{38.99} & \textbf{13.65}  \\
MagPie-Ultra$\overset{+}{}$ 
 + MetaMathQA                  & 44.98 & 5.02 & \textbf{47.08} & 17.56 & 36.77 & 12.18 \\
MagPie-Ultra$\overset{+}{}$ 
 + NuminaMath-CoT                  & 46.27 & \textbf{5.99} & 25.32 & \textbf{18.00} & 37.88 & 12.58 \\
\midrule
\multicolumn{7}{c}{Full SmolTalk}\\
\cmidrule{1-7}
SmolTalk  & 46.67 & 5.49 & 43.75 & 18.60 & 40.02 & 18.19 \\
SmolLM2-SFT$^\dagger$  & \textbf{57.09} & \textbf{6.11} & \textbf{47.54} & \textbf{19.64} & \textbf{42.49} & \textbf{19.06} \\
\bottomrule
\end{tabular}
\end{table}

\clearpage
\section{Long context evaluations}
\label{app:long-context-evaluations}
\begin{figure}[h]
    \centering
    \includegraphics[width=0.8\linewidth]{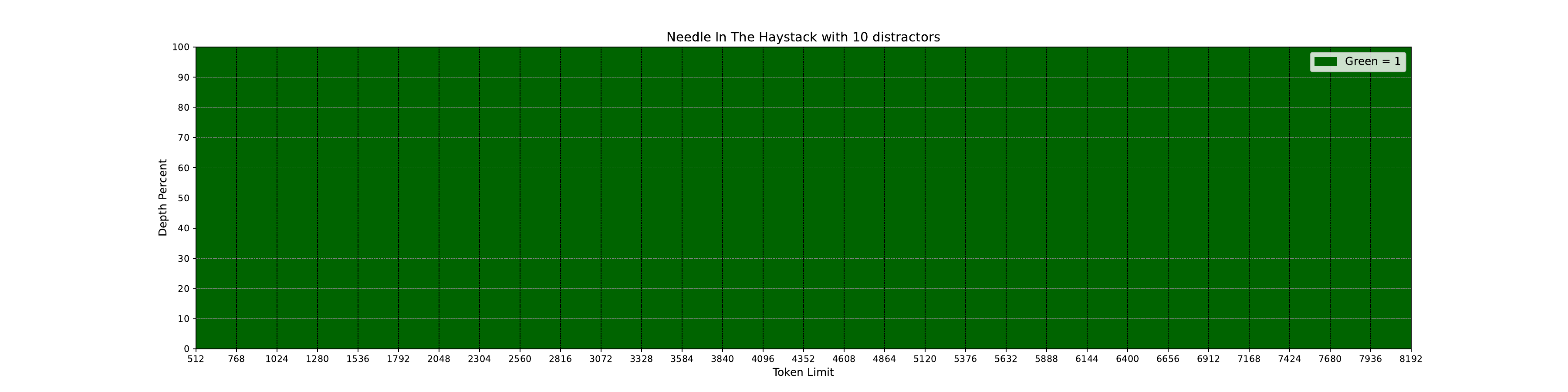}
    \caption{Needle in the Haystack evaluation of SmolLM2 with 8192 context length.}
    \label{fig:niah}
\end{figure}

\cref{fig:niah} shows the evaluation results on the Needle in the Haystack benchmark.

\cref{tab:HELMET} shows the evaluation results on the HELMET benchmark.

\begin{table}[h]
\caption{Evaluation results of the base models on the HELMET benchmark using 8k maximum input length.}
\label{tab:HELMET}
\centering
\begin{tabular}{l r r r}
\toprule
Metric & SmolLM2-1.7B & Llama3.2-1B & Qwen2.5-1.5B \\
\midrule
Average-Real & 31.67 & 35.56 & \textbf{38.76} \\
Average-All & 32.61 & 39.61 & \textbf{44.40} \\
Recall & 36.38 & 55.81 & \textbf{66.94} \\
RAG & 47.17 & 42.13 & \textbf{47.54} \\
ICL & 23.20 & 51.20 & \textbf{52.00} \\
Re-rank & 23.31 & 26.93 & \textbf{29.29} \\
LongQA & \textbf{33.00} & 21.99 & 26.23 \\
\bottomrule
\end{tabular}
\end{table}

\end{document}